\documentclass[10pt,twocolumn,letterpaper]{article}

\usepackage{wacv}
\usepackage{times}
\usepackage{epsfig}
\usepackage{graphicx}
\usepackage{amsmath}
\usepackage{amssymb}
\usepackage{booktabs}

\usepackage{algorithm,algorithmic}
\usepackage{derivative}
\usepackage{soul}

\def\wacvPaperID{1612} 

\wacvalgorithmstrack   

\wacvfinalcopy 

\ifwacvfinal
\usepackage[breaklinks=true,bookmarks=false]{hyperref}
\else
\usepackage[pagebackref=true,breaklinks=true,colorlinks,bookmarks=false]{hyperref}
\fi

\begin{document}

\title{On Quantizing Implicit Neural Representations}

\author{Cameron Gordon\\
University of Adelaide\\
\and Shin-Fang Chng \\ 
University of Adelaide
\and Lachlan MacDonald \\
University of Adelaide \\
\and Simon Lucey\\  
University of Adelaide\\
}

\maketitle
\thispagestyle{empty}

\begin{abstract}
The role of quantization within implicit/coordinate neural networks is still not fully understood. We note that using a canonical fixed quantization scheme during training produces poor performance at low-rates due to the network weight distributions changing over the course of training. In this work, we show that a non-uniform quantization of neural weights can lead to significant improvements. Specifically, we demonstrate that a clustered quantization enables improved reconstruction. Finally, by characterising a trade-off between quantization and network capacity, we demonstrate that it is possible (while memory inefficient) to reconstruct signals using binary neural networks. We demonstrate our findings experimentally on 2D image reconstruction and 3D radiance fields; and show that simple quantization methods and architecture search can achieve compression of NeRF to less than 16kb with minimal loss in performance (323x smaller than the original NeRF). 

\end{abstract}

\section{Introduction}
\label{sec:intro}


There is increasing interest in the compression of implicit neural functions \cite{dupont_coin_2021,dupont_coin_2022,strumpler_implicit_2021,xie_neural_2022}. While existing works have examined the use of quantization as part of a neural compression pipeline, there remain a number of classical quantization methods that have been less applied to these problems \cite{gholami_survey_2021,gray_quantization_1998,gersho_vector_1992}. In particular, within compression of implicit neural functions the usual method is to apply \textit{uniform} quantization \cite{dupont_coin_2021,dupont_coin_2022,strumpler_implicit_2021}, and to use a \textit{fixed} quantization scheme which does not change over the course of training. While simple and efficient, this can introduce quantization error if the underlying distribution varies across training. 

\begin{figure}
    \centering
    \includegraphics[width=0.45\textwidth]{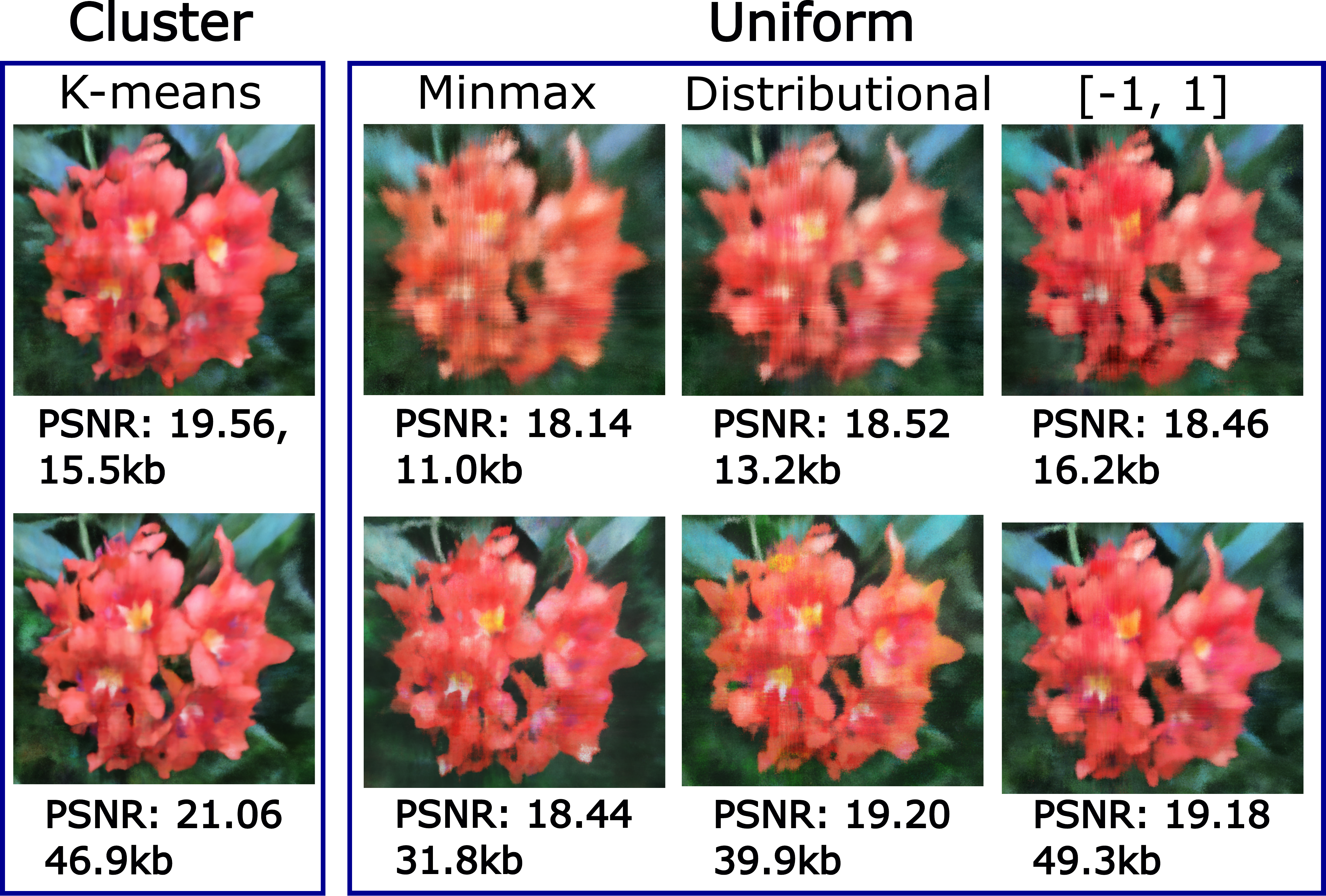}
    \caption{Comparison of cluster and uniform quantization on a small NeRF model. Top: 4-layers, 64 neurons per layer. Bottom: 4-layers, 128 neurons per layer. Quantization to 3-bits-per-weight.}
    \label{fig:3bitflower}
\end{figure}

In this work, we apply a cluster quantization method to more closely represent the weights of implicit neural representations. In quantization literature a key idea is to match the \textit{distribution} of the quantized and original signals as closely as possible to prevent reconstruction error; this is achievable through the use of clustered partitioning of signals \cite{gersho_vector_1992,gray_quantization_1998}. In addition, it is well-known that the distribution of weight values in a network changes over the course of training - as such the distributional assumption at one epoch - may not be valid over the entire training cycle. 

Furthermore while it is known that uniform quantization methods can enable representation of signals with a high degree of fidelity, the trade-off between network capacity and the level of feasible quantization has been less explored \cite{dupont_coin_2021,dupont_coin_2022, strumpler_implicit_2021}. Intuitively it would ordinarily be expected that increasing the amount of network quantization should be beneficial in reducing the rate (i.e. size in bits) of the network. However, to maintain the same reconstruction quality one needs to dramatically increase the size of the network to offset the loss of fidelity in the weights. Surprisingly, we show that a \textit{higher} quantization level with a \textit{restricted} network structure can be more memory efficient than a \textit{low} quantization level with a more expressive network structure. 

The broad focus of this paper is a comparative analysis of cluster and uniform quantization in implicit neural representations at low bits-per-weight. In particular, our contributions are as follows:
\begin{itemize}
    \item We introduce an adaptive clustering strategy for quantization-aware training applied to implicit neural networks, showing improved performance at lower quantization levels than uniform methods for multiple modalities (images and neural radiance fields).
    \item We demonstrate that a performance trade-off between the level of quantization and the expressivity of network architectures is required to adequately reconstruct a signal, with high-fidelity reconstructions able to be achieved even with a binary quantization.
    \item As an application of the analysis, we demonstrate that substantial compression of neural radiance fields (323x smaller than the original NeRF \cite{mildenhall_nerf_2020} and 58x smaller than cNeRF \cite{bird_3d_2021}) can be obtained with even simple quantization methods and architecture search with minimal collapse in performance. 
\end{itemize}

\section{Background} 

\subsection{Quantization} 

\begin{figure}
    \centering
    \includegraphics[width=0.45\textwidth]{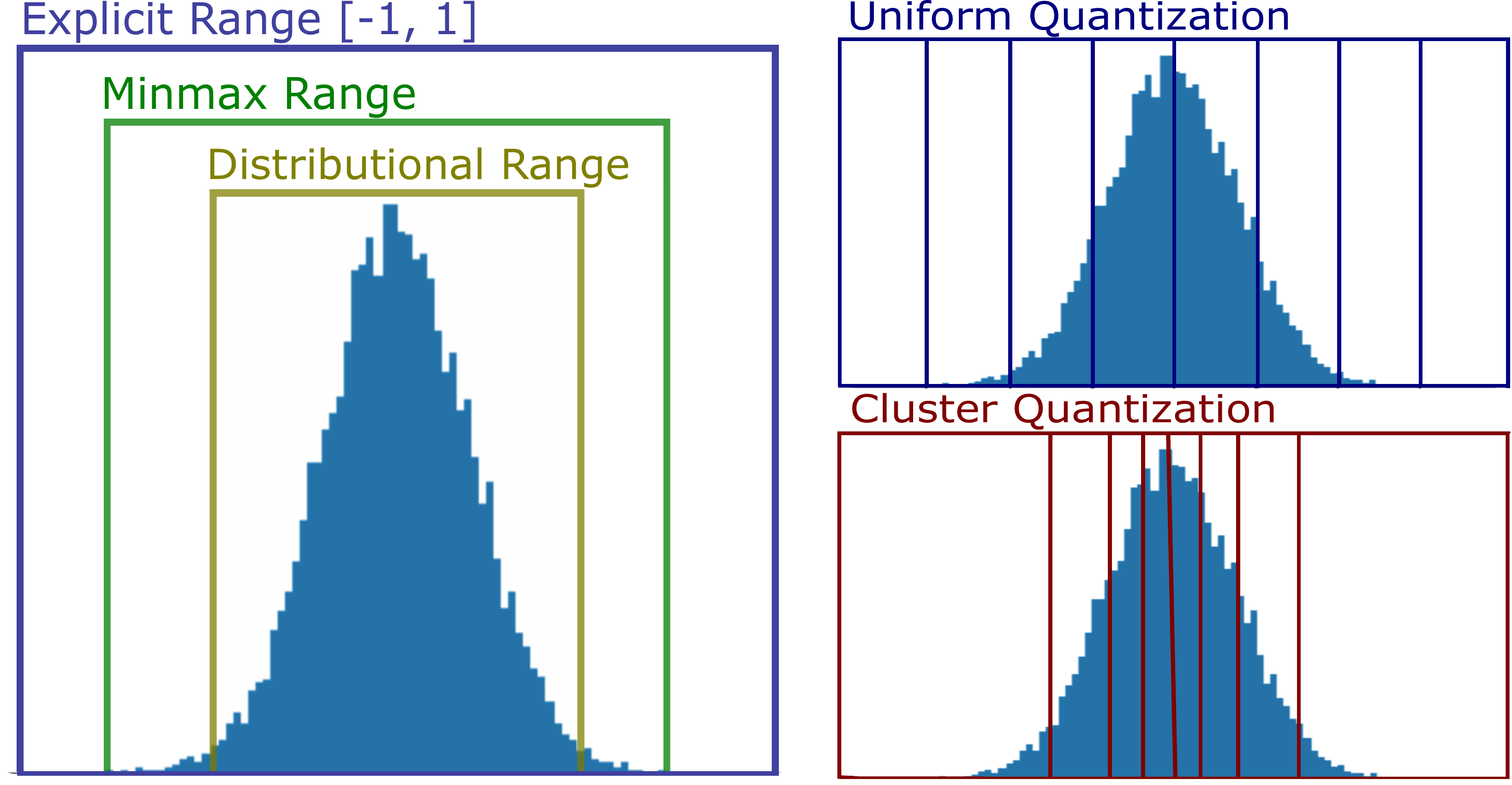}
    \caption{Left: Different uniform quantization ranges. Right: Comparison between decision boundaries for the examined uniform and cluster quantizations. A uniform quantization scheme equally divides the quantization range. A cluster quantization divides the quantization range such that each partition contains equal mass of the distribution.}
    \label{fig:distr}
\end{figure}

We define a quantization scheme $Q: \mathcal{A} \to \mathcal{B}$ as a map between two sets $\mathcal{A}, \mathcal{B}$ such that the cardinalities $|\mathcal{A}| \geq |\mathcal{B}|$ (\eg $Q: \mathbb{R} \to \mathbb{Z}_5$). While other alphabets are possible, typically the codomain has elements in $\mathbb{R}$ or $\mathbb{Z}$. For example, a uniform quantization scheme has a codomain with an equal spacing within the range $(a,b)$. In contrast a non-uniform quantization scheme has non-equal spacing between its elements. Examples of non-uniform quantization schemes include \textit{logarithmic} quantization, in which elements are logarithmically spaced; and cluster quantization, in which elements are partitioned by \textit{decision boundaries} separating clusters of points \cite{gholami_survey_2021,gray_quantization_1998,gersho_vector_1992}. The Lloyd Algorithm and K-means are examples of a cluster quantization in which decision boundaries are determined to have equal data mass, with the values in each partition mapped to the \textit{centroid} (mass centre) of the partition \cite{gersho_vector_1992,gray_quantization_1998}. 

A \textit{data-dependent} quantization scheme can be defined as a quantization map determined with respect to the data to be quantized. A \textit{data-agnostic} quantization scheme has a mapping determined a priori. For a uniform quantization scheme a data-dependent approach may be to set the range to be the (min, max) values of the data, or be a clipped range incorporating distributional information such as standard deviations of the data \cite{gray_quantization_1998,gersho_vector_1992}. Figure \ref{fig:distr} shows the difference between these approaches. A \textit{fixed} quantization scheme uses the same quantization scheme for every epoch $t$. That is, $Q_t = Q_{t+1}; \forall t$. An \textit{adaptive} quantization scheme allows $Q_t$ to vary over the course of training. 

The \textit{quantization error} of a scalar $x$ is given by \cite{gersho_vector_1992}: 
\begin{equation}
    \epsilon = x - Q(x).
\end{equation}

For a matrix of values we can use the L2-norm of quantization errors as a distance metric: 
\begin{equation}
e = \|X -  Q(X) \|^2_2.
\end{equation} 

For an L-layer perceptron we can define the \textit{total layer-wise quantization error} (TLQE) to be given by: 
\begin{equation}\label{TQE} 
    TLQE = \sum_{l \in L} \|W_l - Q_l(W_l) \|^2_2,
\end{equation}
where $W_l$ refers to the full precision weights at layer $l$ and $Q_l$ is the quantization mapping applied to layer $l$. 


\subsubsection{Fixed and Adaptive Quantization}

Consider $Q_t$ as depending on the weights at a given epoch $t$. Under a fixed data-dependent cluster quantization, we have $Q_t$ defined as a mapping that minimises the distance between quantized and unquantized values at epoch $t$: 
\begin{equation}\label{Q}
    Q_t := \underset{Q}{\text{argmin}} \|W_{t}- Q(W_{t})\|^2_2.
\end{equation} 

Note that typically the distribution of weight values changes over training. Consider applying the same $Q_t$ to a matrix of weights at time $t'$ after several epochs of training has occurred. By assumption from Equation \ref{Q}, we have that $Q_t$ is the optimal mapping for epoch $t$ and $Q_{t'}$ the optimal mapping for epoch $t'$. As a result, we know that the quantization error that occurs from using $Q_{t}$ at $t'$ is greater than or equal to that of applying an optimal $Q_{t'}$:
\begin{equation}
    \|W_{t'}-Q_t(W_{t'}) \|^2_2 \geq \|W_{t'}-Q_{t'}(W_{t'}) \|^2_2.
\end{equation}
This motivates the use of an adaptive quantization scheme. As repartitioning with a K-means algorithm is computationally costly it is possible to set an adaptive quantization rule based on an interval of epochs, or based on the quantization error such that repartition occurs if for a chosen $\delta$: 
\begin{equation}
    \|W_{t'}- Q_{t}(W_{t'})\|^2_2 \geq \|W_{t}- Q_t(W_{t})\|^2_2 + \delta. 
\end{equation}


In practice, we find it sufficient to repartition periodically based only on the number of epochs.


\subsection{Implicit Neural Representations (INR)}
An INR is a function mapping a coordinate input vector $\textbf{x}$ to an output feature vector $\textbf{y}$ parameterised by neural network weights \cite{sitzmann_implicit_2020,dupont_coin_2021,strumpler_implicit_2021,xie_neural_2022} as

\begin{equation}
    f_\theta(\textbf{x}) \to \textbf{y}.
\end{equation}
Examples of INRs include coordinate networks \cite{ramasinghe_beyond_2022,ramasinghe_learning_2021}, NeRF and its many variants \cite{tewari_advances_2022,xie_neural_2022,mentzer_high-fidelity_2020,park_nerfies_2021}, audio \cite{sitzmann_implicit_2020,dupont_coin_2022}, video \cite{chen_nerv_2021,zhang_implicit_2021,le_mobilecodec_2022}, topological representations \cite{zehnder_ntopo_2021}, light-field representations \cite{feng_signet_2021}, implicit geometry \cite{darmon_improving_2021, palmer_deepcurrents_2022}, novel view synthesis, volumetric scalar fields \cite{lu_compressive_2021}, and gigapixel image fitting \cite{martel_acorn_2021}. A special case of INRs are image regression problems \cite{sitzmann_implicit_2020,strumpler_implicit_2021,dupont_coin_2021}. An image regression learns a representation function mapping: $f_\theta(x,y) \to (r,g,b)$ for a single image. The network weights $\theta$ provide an encoding that predicts the approximate pixel value for a given coordinate. A forward pass across the original set of coordinates approximately reconstructs the original image. A sufficiently small or quantized network can therefore be treated as a form of lossy image compression \cite{strumpler_implicit_2021,dupont_coin_2021,dupont_coin_2022}.

\subsection{Neural Radiance Fields (NeRF)} 
A NeRF is an implicit neural representation of the form
\begin{equation}
    f_\theta(x,y,z,\theta, \phi) \to (r,g,b,\sigma),
\end{equation} 
where spatial location is provided by the coordinates ($x, y, z$) and viewing direction ($\theta, \phi$) \cite{mildenhall_nerf_2020}. When trained on a set of camera poses for a given scene, the implicit representation enables the interpolation and generation of novel pose estimates. A wide number of technical variations have extended the original model for purposes of improved fidelity or rendering speed, such as \cite{reiser_kilonerf_2021},~\cite{lin_barf_2021} and~\cite{chng_garf_2022}. While the implicit nature of NeRF is itself a compression form, only a few works have investigated further compressing this model. Methods have included a mixture of rank-residual decomposition, quantization, entropy-penalisation, pruning, and distillation techniques \cite{isik_neural_2021,bird_3d_2021,takikawa_variable_2022,tang_compressible-composable_2022,shi_distilled_2022}. 

\subsection{Related Works}

\subsubsection{Quantization in Deep Learning}

How to quantize a signal to preserve its relevant information content has been of fundamental interest in signal processing, information theory, compression, and other fields since at least the time of Shannon \cite{shannon_communication_1949}. Within image processing, quantization is a fundamental element of image compression algorithms such as JPEG \cite{salomon_data_2007, szeliski_computer_2022}. While the term quantization can refer to the discretisation of any continuous signal, a distinction should be made to its most common usage in computer vision: the reduced-precision quantization involved in representing a floating point value in a reduced number of bits \cite{cover_elements_2006,szeliski_computer_2021}. Within deep learning, reduced-precision quantization has been widely applied to the \textit{weight} and/or \textit{activation} values in a deep feedforward network \cite{han_deep_2016,rastegari_xnor-net_2016,gholami_survey_2021,qin_binary_2020,jacob_quantization_2017}. As deep learning libraries such as PyTorch and Tensorflow typically represent weights in 32-bit or 64-bit format, a lower-precision quantization can lead to memory and speed improvements. At the extreme case, this involves \textit{binary} neural networks whose weights are quantized to take binary values of \{-1, 1\} \cite{courbariaux_binarized_2016}. This was examined in Rastegari \etal \cite{rastegari_xnor-net_2016} which introduces a method of training full-precision networks which is robust to a quantization transformation, known as \textit{quantization-aware training}. A variety of quantization methods for deep neural networks including non-uniform mappings, global and layer-wise mappings, integer quantization \cite{jacob_quantization_2017}, and mixed-precision quantization are also widely described in the literature \cite{han_deep_2016,gholami_survey_2021}.





\subsubsection{Quantizing Implicit Neural Representations}
Within the wider INR literature, several works have examined the use of quantization for compression. Works related to ours are Dupont \etal \cite{dupont_coin_2021} which applied quantization and architecture search to compress images, Strumpler \etal \cite{strumpler_implicit_2021} which applied quantization-aware training and entropy coding, and Chiarlo \cite{chiarlo_implicit_2021}, which looked at compression methods for INRs (distillation, pruning, quantization, and quantization-aware training). Dupont \etal \cite{dupont_coin_2022} showed impressive results compressing INRs over multiple modalities through the use of quantized weight modulations. Each of these papers apply uniform rather than cluster quantization. Furthermore each of recent \cite{strumpler_implicit_2021}, \cite{dupont_coin_2022}, and \cite{lee_meta-learning_2021} apply a meta-learned preinitialisation (MAML) to improve rate-distortion performance and model performance. They require an additional dataset for the preinitialisation; in contrast we learn on single signal instances to directly compare quantization methods. 

In terms of works employing cluster quantization to implicit neural representations, we find Lu \etal \cite{lu_compressive_2021}, Takikawa \etal \cite{takikawa_variable_2022}, and Shi \etal \cite{shi_distilled_2022} to be the most similar to our method. Lu \etal \cite{lu_compressive_2021} applied a K-means clustering to the weights of each layer of a network to compress volumetric scalar fields. 
We differ from their work by investigating NeRF and 2D image compression; and in addition employ a quantization-aware training and entropy compression. The second is the recently released Takikawa \etal \cite{takikawa_variable_2022}, who applied a learned quantization to feature grids to compress NeRF and signed-distance fields. While they used K-means clustering as a post-processing benchmark comparison, they did not use it as part of quantization-aware training as in our method. More recently, Shi \etal \cite{shi_distilled_2022} applied a low-rank decomposition and distillation to a trained model based on the original NeRF (8 layers, 256 channels), before a global K-means quantization map is used iteratively to quantize each layer with other layers then retrained. In contrast, we apply layer-wise quantization-aware training throughout our entire training procedure. Furthermore, our work directly investigates the impact of quantization methods, network architectures, and quantization levels have on compression across different modalities. 


\section{Method}
For a loss function $L$, number of hidden layers $h$, number of hidden units per layer $w$, quantization function $q$, bits-per-weight $k$, compression function $C$ and target memory constraint $D$, we can view the compression of implicit neural representations through as a constrained optimization across architectures and quantization levels as
\begin{align}
    & \min \qquad L(\cdot)  \\
    &\text{s.t.} \qquad C(q, w, h, k) \leq D. \nonumber
\end{align}




\subsection{Quantization Methods}

 Our experiments are performed on 4 quantization methods (\textit{Explicit [-1, 1]}, \textit{Distributional}, \textit{Minmax}, and \textit{K-means}). The first three are uniform quantizations with different ranges (see Figure \ref{fig:distr}): an explicit range between $[-1,1]$; a \textit{distributional} range calculated as a function of the standard deviation of the weight distribution; and a \textit{Minmax} quantization range determined by the minimum and maximum value of the distribution. These are compared with a clustered calculated using the K-means algorithm. 
 
 
The Explicit [-1, 1] uniform quantization is calculated using a $k$-bit formula found in Rastegari \etal \cite{rastegari_xnor-net_2016}:
\begin{equation}\label{rastegari}
    q_k(x) = 2(\frac{\text{round}((2^k-1)(\frac{x+1}{2}))}{2^k-1}-\frac{1}{2}), 
\end{equation}
where $x \in [ -1, 1 ]$.  
The Distributional quantization uniformly quantizes within $d$ standard deviations of the weight distribution, where $d$ is calculated according to a $k$-bit formula found in Dupont \etal \cite{dupont_coin_2022}: 
\begin{equation}\label{dupont_quant}
d = 3 + \frac{3(\text{k}-1)}{15}.
\end{equation}
\begin{figure}[b]
    \centering
    \includegraphics[width=0.3\textwidth]{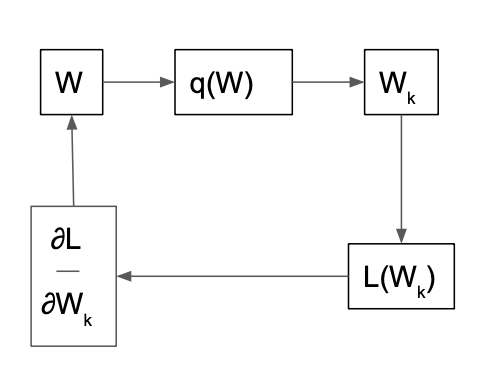} 
    \caption{QAT Per epoch training cycle: The current full-precision weight matrix $W$ is quantized $q(W)=W_k$. The loss function $L(W_k)$ is calculated for the input data, and the error derivatives $\pdv{L(W_k)}{W_k}$ calculated. The full-precision weight matrix $W$ is then updated using backpropagation.}
    \label{fig:qatraining}
    
\end{figure}

\begin{algorithm}[t]
\textbf{Input:} {bits-per-weight $k$, model, repartition\_epoch}
\begin{algorithmic}[1]

\FOR{epoch $1$ to $N$}
\STATE Train model (quantization-aware training)
\IF{repartition\_epoch}
\FOR{$l$ in layers} 
\STATE Recalculate quantization map (1D K-means)

\ENDFOR
\ENDIF
\ENDFOR
\STATE Convert Quantized Model to Codebook 
\STATE BZIP2 Codebook Model + Cluster Dictionary  
\end{algorithmic}
\caption{Adaptive Cluster Quantization}
\label{alg:alg_adaptive}
\end{algorithm}

\subsection{Quantization-Aware Training (QAT)} 
\begin{figure}[t]\label{repartitioning}
    \centering
    \includegraphics[width=0.45\textwidth]{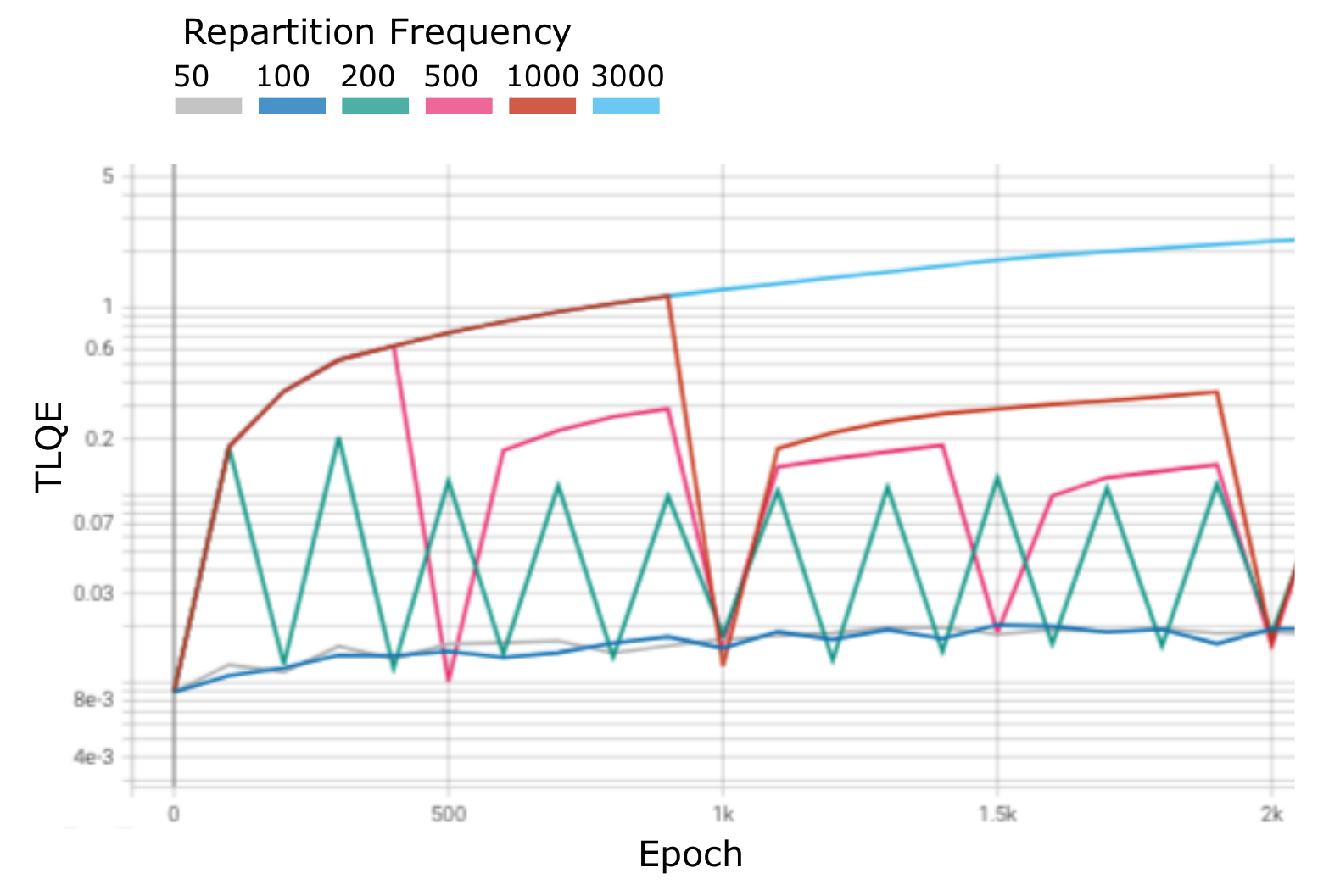}
    \caption{Effect of repartitioning on Total Layerwise Quantization Error (TLQE) [log scale]. Recalculating partitions reduces quantization error at the cost of increased computation. Architecture: 1 hidden layer, 18 neurons, 5-bit weights, K-means quantization. }
    \label{fig:repartition}
\end{figure}

We employ the quantization-aware training introduced by Rastegari \etal \cite{rastegari_xnor-net_2016}. The algorithm is briefly recalled in Figure \ref{fig:qatraining}. This procedure trains the model robustly to quantization through the use of a straight-through estimator \cite{bengio_estimating_2013, rastegari_xnor-net_2016}. We adapt this procedure to include \textit{periodic repartitioning} in which the quantization mappings for the Distributional, Minmax, and K-means are recalculated periodically over training. The Explicit $[-1 , 1]$ quantization is fixed for all epochs. Repartitioning every $F$ epochs in a $N$ epoch training cycle introduces a computational overhead of $\frac{NG(q,w,h,k)}{F}$, where $G$ is a cost of repartitioning which depends on the network architecture ($w,h$), quantization function ($q$), and bits-per-weight ($k$). Periodic repartitioning therefore strikes a balance between accumulating quantization error and the introduced computational overhead of recomputing partitions as weight distributions change over training; see Figure \ref{fig:repartition} and Sec.~\ref{sec:dis}. Following training, the quantized weights are converted to an integer representation (a dictionary mapping of the quantized float and the integer) as \cite{strumpler_implicit_2021}. Both the integer representation and the mapping are then compressed using BZIP2, an entropy encoding compression library. Algorithm \ref{alg:alg_adaptive} describes our approach.

\subsection{Implementation Details}\label{subsec:impl}
Training was conducted with the Adam optimizer with the hyperparameters $1e^{-4}, \beta=(0.99, 0.999)$ and weight decay $=1e^{-8}$ \cite{kingma_adam_2014}. For image regression we use a sine activation with frequency 30 as described \etal \cite{dupont_coin_2021,strumpler_implicit_2021}, but compare with Gaussian and ReLU activations as supplemental ablation \cite{dupont_coin_2021,sitzmann_implicit_2020}. For our 2D image regression experiments, we experimented with both the MSE loss and its negative base-10 logarithm (i.e. an unscaled peak signal-to-noise ratio (PSNR)). For the NeRF experiments, we use the MSE loss \cite{mildenhall_nerf_2020}; see Sec.~\ref{sec:dis} for more details. Our experiments are evaluated quantitatively using standard perceptual metrics, including the PSNR, the structural similarity index measure (SSIM), and LPIPS Alex and VGG (two learned perceptual metrics) \cite{wang_image_2004,zhang_unreasonable_2018,salomon_data_2007,szeliski_computer_2022,gersho_vector_1992}. We additionally evaluate the \emph{gradient PSNR}, which we define by
\begin{equation}
    PSNR_\Delta = -\log_{10}(\|\Delta(f_\theta(x,y))-\Delta(X) \|^2_2),
\end{equation}
where $\Delta(\cdot)$ is an approximation of the image gradient generated through the Sobel operator \cite{szeliski_computer_2022}.  The gradient PSNR is used to determine the quality of preserved image gradients, as these are often important for downstream tasks such as image classification and segmentation \cite{szeliski_computer_2022}.

\begin{figure}[b]
    \centering
    \includegraphics[width=0.45\textwidth]{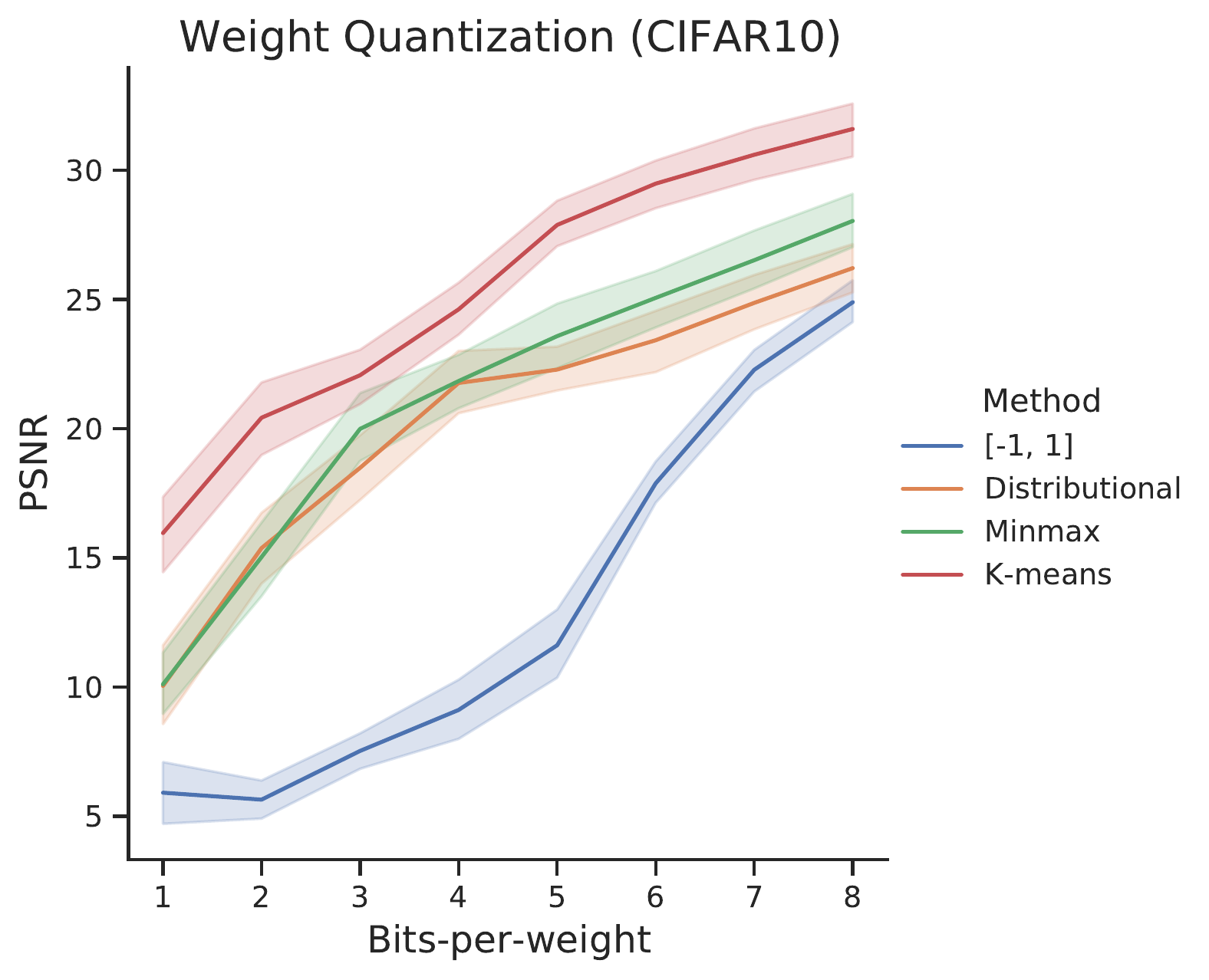}\\

    \includegraphics[width=0.45\textwidth]{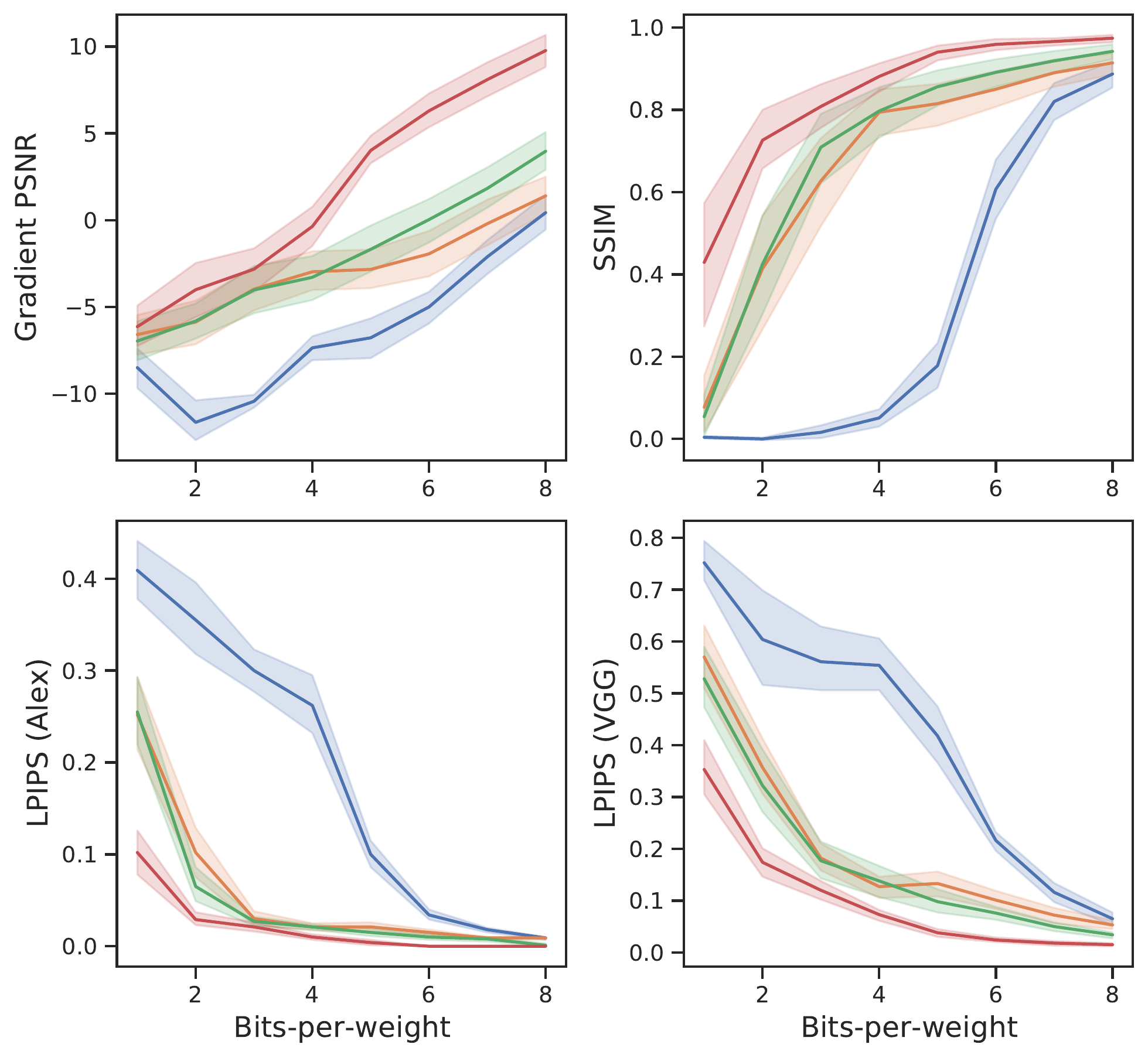}

    \caption{Perceptual evaluation (CIFAR10)}
    \label{fig:cifarperceptual}
\end{figure}





\section{Results} In this section we present our comparison of different quantization schemes. First, we demonstrate a performance difference between cluster and uniform quantization for image regression of the CIFAR10 Dataset. This is followed by a deeper architectural analysis on an instance of the DIV2K Dataset. As an example application of the analysis we apply our method to the more complicated modality of NeRF.


\subsection{CIFAR10}
Experiments for 2D image reconstruction were conducted on the CIFAR10 dataset. We use a base network of 1 hidden layer, sine activation, 20 neuron layers replicating the architecture used for COIN compression experiments in \cite{dupont_coin_2022}. Repartitioning was applied at every epoch. As can be seen in Figure \ref{fig:cifarperceptual}, the K-means cluster quantization has improved reconstruction relative to the other methods as evaluated across perceptual metrics. Interestingly, it it found to have substantially higher accuracy according to the Gradient PSNR even at high bits-per-weight. Comparing the uniform approaches, Distributional and Minmax quantization gives broadly similar results. Quantizing uniformly between a fixed [-1, 1] gives the worst performance with little signal reconstructed at low bits-per-weight.  




\subsection{DIV2K}
\begin{figure}[h!]
    \centering
    \includegraphics[width=0.45\textwidth]{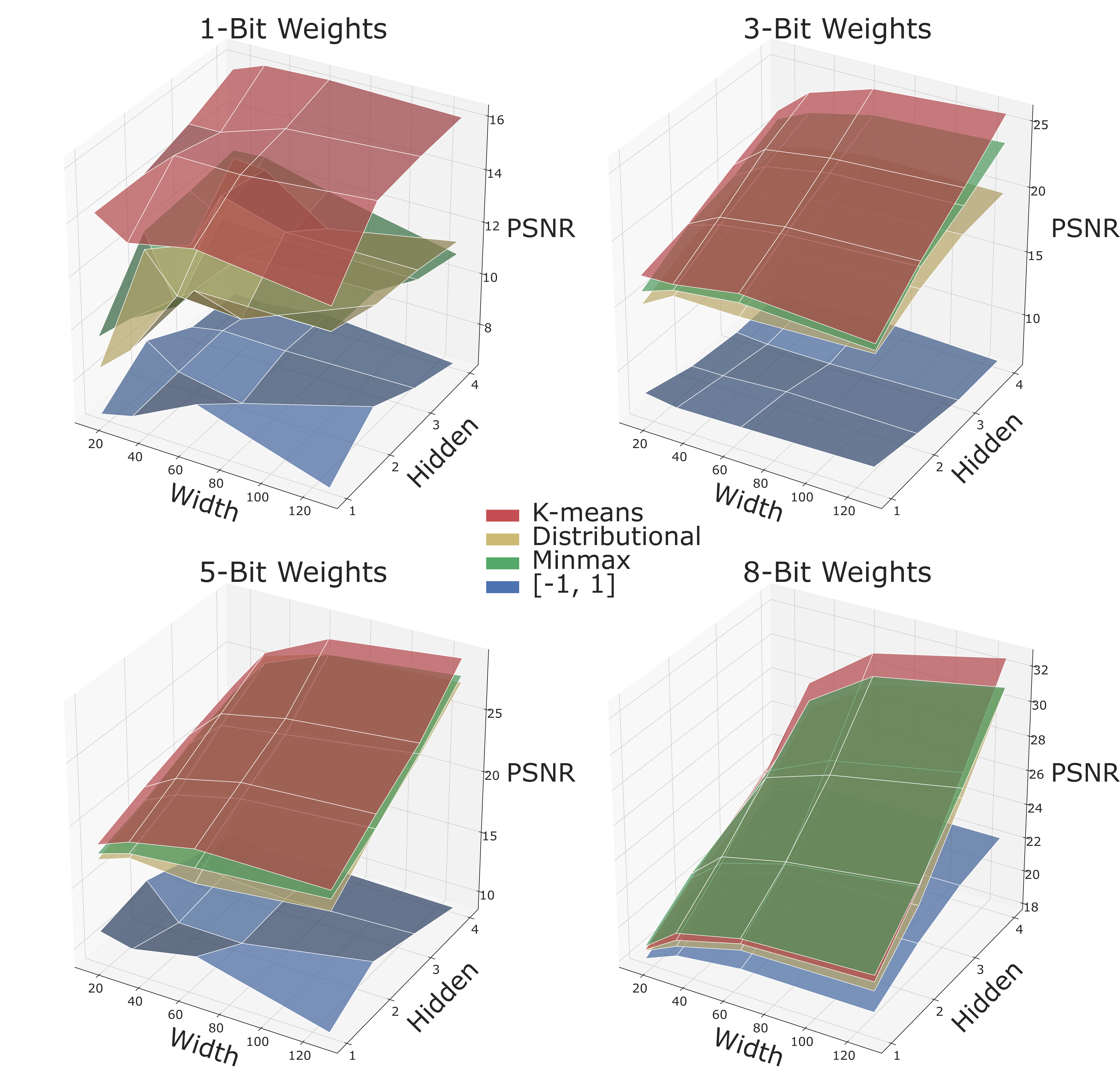}
    \caption{Comparison between quantization methods under different architectures at different rates on DIV2K index 3. The clustered quantization outperforms uniform methods at low bits-per-weight and the difference becomes marginal at higher resolutions.} 
    \label{fig:quantcomparediv2k}
\end{figure}

We conducted the experiments for 2D image reconstructions on the DIV2K image test suite, which consists images with at least 2000 pixels on at least one axis. Figure \ref{fig:quantcomparediv2k} compares the quantization methods at different architectures and rates of quantization. Repartitioning was applied every 50 epochs. As expected, the K-means quantization outperforms the uniform quantization methods at low bit-rates. At 8 bits-per-weight the difference between K-means, Distributional, and Minmax quantization is negligible. This is consistent with the well-known result that high resolution uniform scalar quantizers closely approximate an optimal quantization (to within 2.82dB for a Gaussian source), a result first noted by Koshelev in 1963 and since rediscovered by multiple authors \cite{gray_quantization_1998,gersho_vector_1992}.



\begin{figure}[h!]
    \centering
    \includegraphics[width=0.21\textwidth]{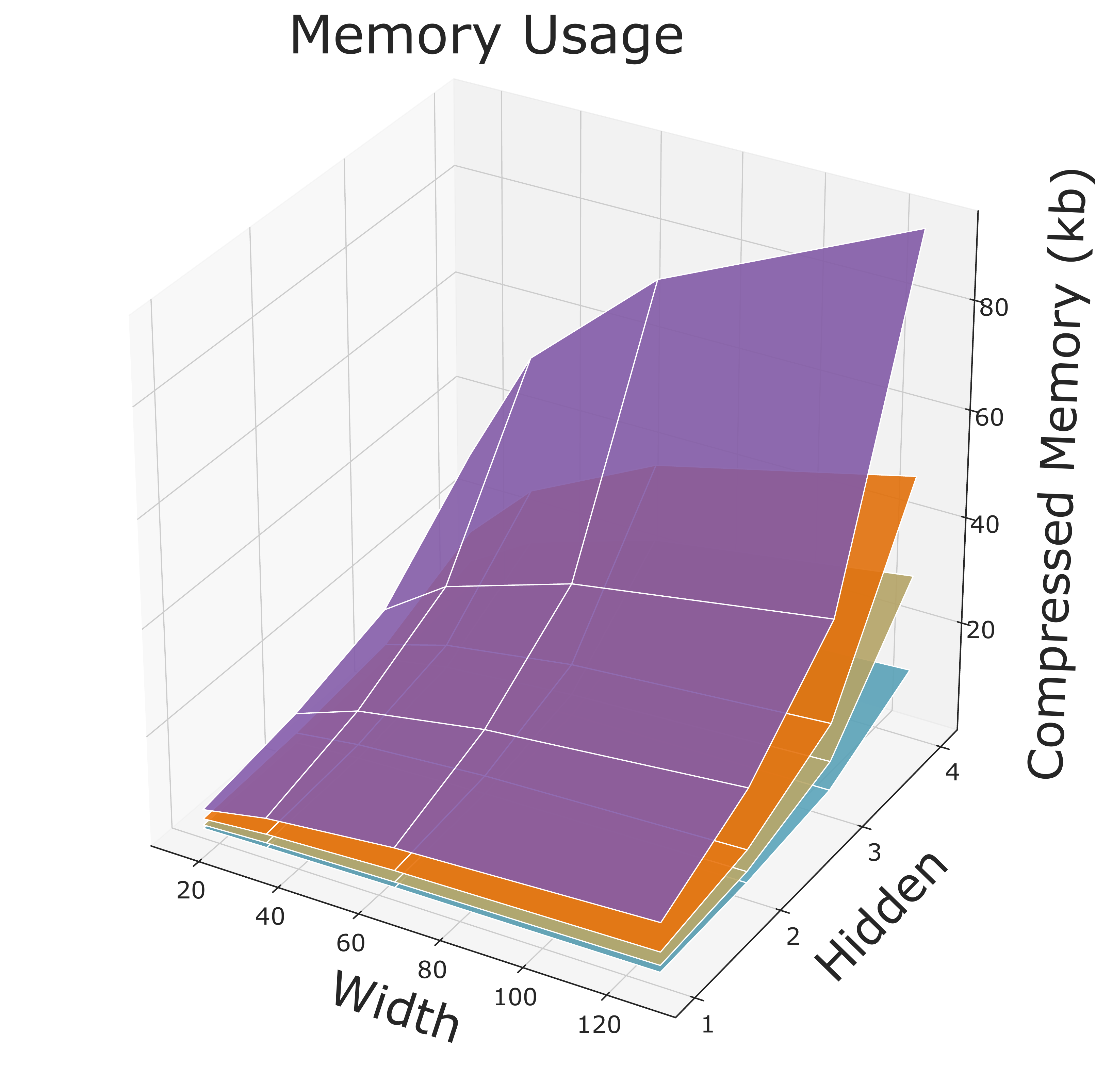}~
    \includegraphics[width=0.270\textwidth]{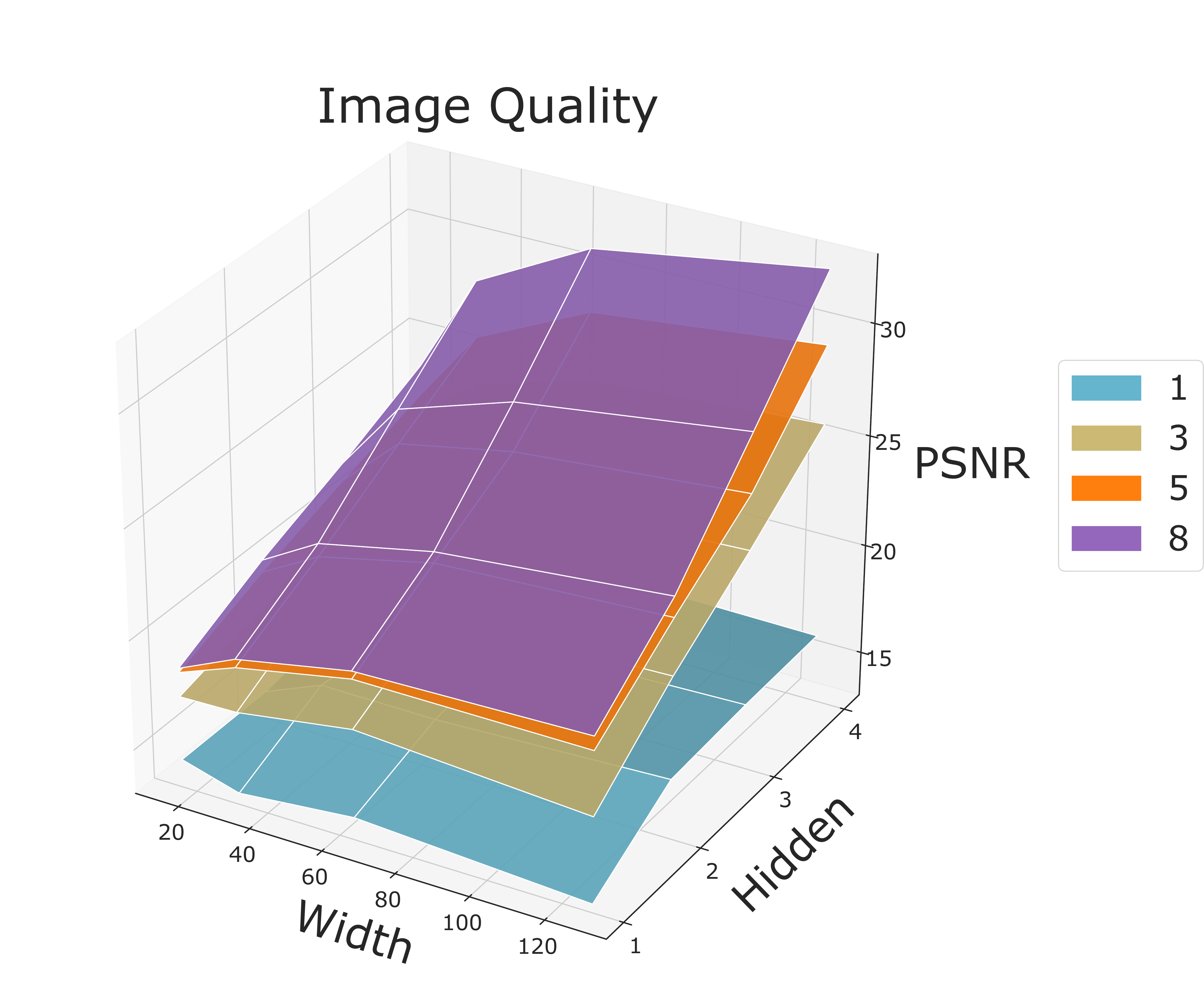}
    
    \caption{Architecture search over the number of hidden layers, and the number of units per hidden layer. The colours correspond to different quantization bits-per-weight (1, 3, 5, 8) using K-means quantization. Div2K index 3, evaluated at 3000 epochs.}
    \label{fig:div2ksurface}
\end{figure}


\subsection{NeRF Experiments} 

For evaluations on neural radiance fields, we used a 4-layer NeRF~\cite{mildenhall_nerf_2020} with 64 hidden units per layer without hierarchical sampling trained for 200,000 epochs with repartitioning every 100 epochs. The highest performing epoch was selected for evaluation. A ReLU activation with positional encoding is used as per ~\cite{mildenhall_nerf_2020}. Note that our analysis is agnostic to activations; see supplemental ablation. Initial experiments were performed on the LLFF `flower' instance with bits-per-weight (1, 3, 5, 8). Following this an evaluation was conducted on the full LLFF and Blender Datasets using with weights quantized to 3 bits-per-weight. 

\begin{figure*}
    \centering
    \includegraphics[ width=0.85\textwidth]{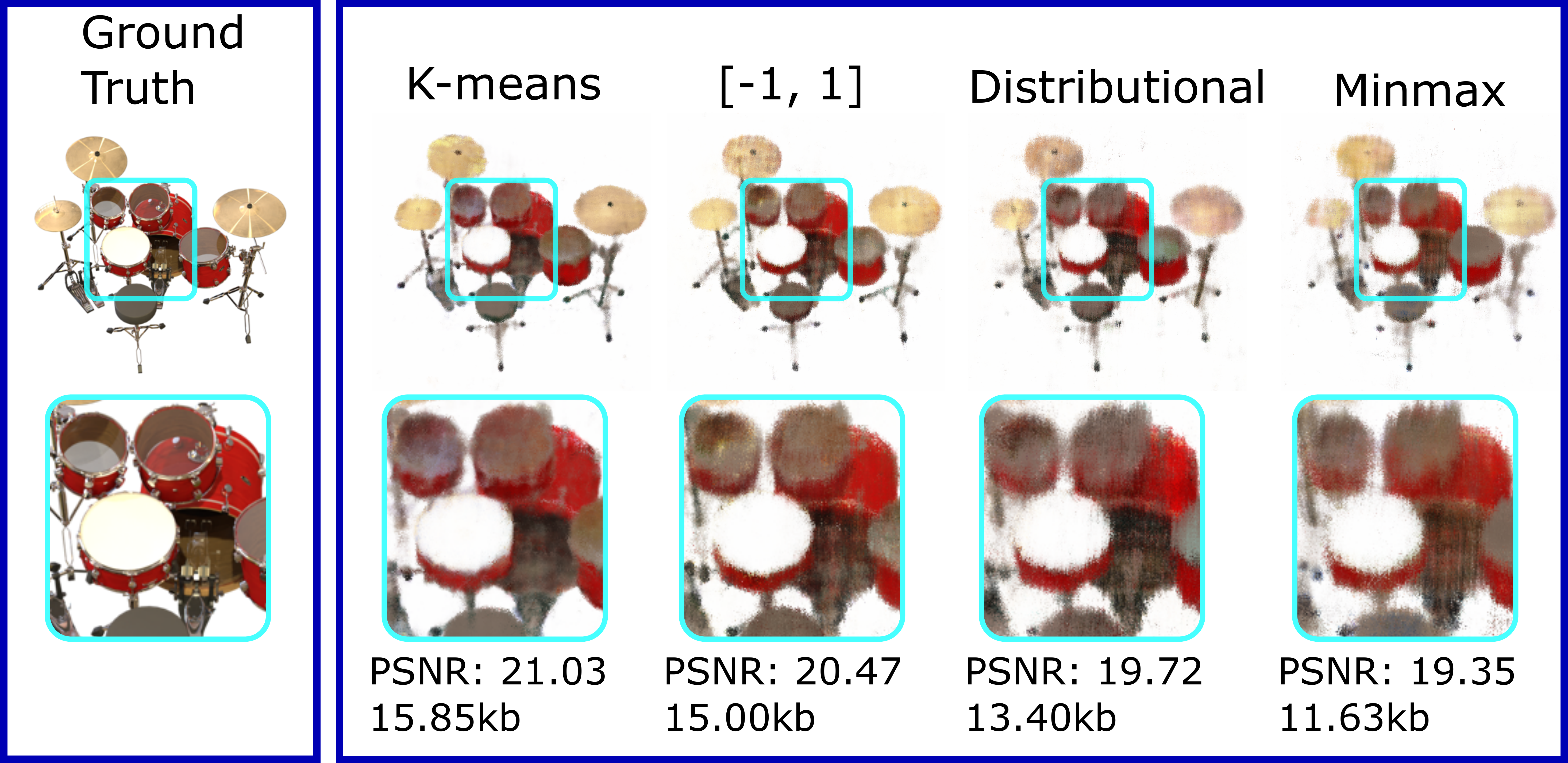}
    \caption{Synthetic NeRF Qualitative Result on ``drum'' instance. Less distortion is visible under the K-means quantization. Each model is less than 16kb (4 hidden layers of 64 neurons, with 3-bit quantization).}
    \label{fig:drums}
\end{figure*}

\begin{figure} 
    \centering
    \includegraphics[width=0.45\textwidth]{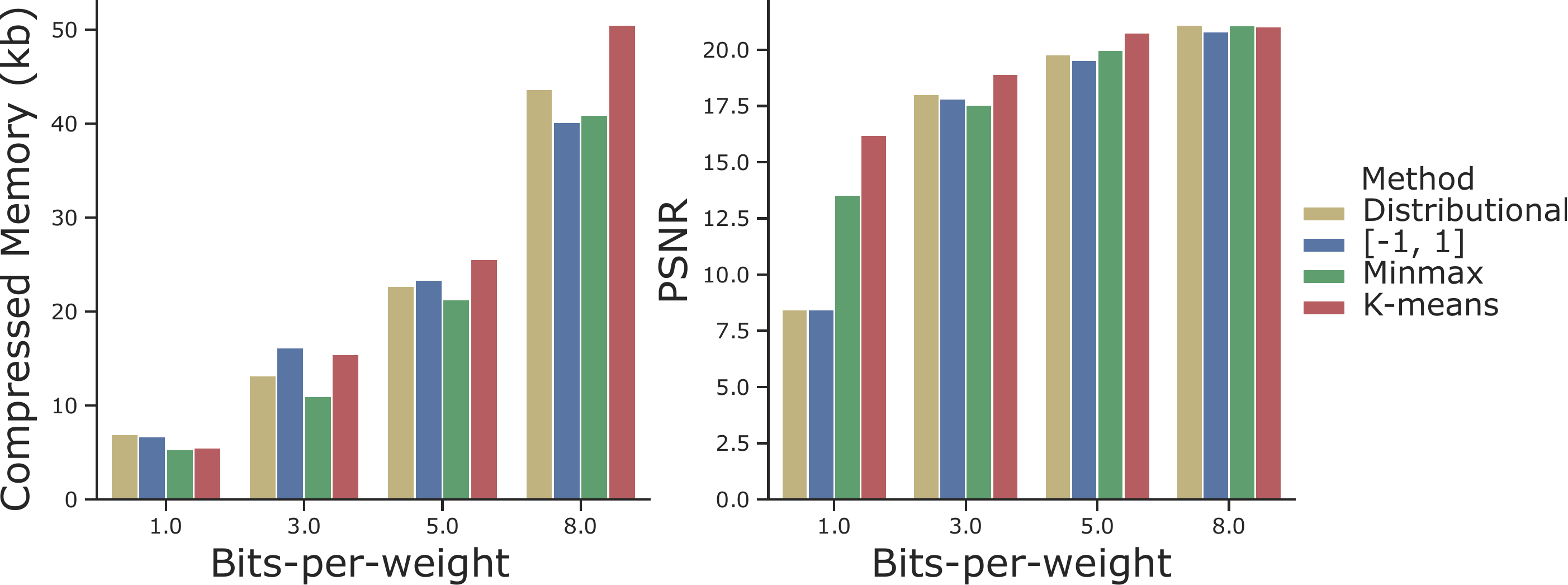}
    \caption{Comparison of PSNR and Compressed Memory Size for NeRF Flower images. Architecture (4-layers, 64 neuron layers).}
    \label{fig:NeRF}
\end{figure}

\paragraph{LLFF Dataset}
Figure \ref{fig:NeRF} shows our results in compressing the NeRF on a `flower' instance (LLFF) using uniform and cluster quantization. Significant compression of NeRF is observed without a catastrophic degradation of accuracy (\eg 20.77 PSNR with a model size of 25.6kb). In terms of memory usage, this compares favourably to that obtained by both the original NeRF (27.42 PSNR, 5169kb) and cNeRF (27.39 PSNR at 938kb) \cite{mildenhall_nerf_2020,bird_3d_2021}. Furthermore, we note the improvement of K-means quantization at low bits-per-weight. At 3 bits-per-weight K-means quantization obtains a test PSNR of 18.93 with an architecture of 4-layers 64 neurons, compared to 17.82 under Explicit [-1,1] quantization, 17.56 under Minmax quantization, and 18.02 under a Distributional quantization. Moreover K-means quantization enables the signal to be obtained under 1-bit quantization (16.21 PSNR); a restriction that causes the signal to collapse under both Explicit and Distributional quantization. At higher bits-per-weight this benefit is diminished. This is expected and consistent with quantization theory, as increasing partitions reduces the uncovered distributional support \cite{gersho_vector_1992,gray_quantization_1998}. Architecture choice has a large impact on memory footprint (\eg increasing hidden layer neurons to 128 approximately triples the used memory from 25.6kb to 76.9kb), with a positive but marginal improvement in PSNR (20.77 to 21.61; see Supplementary Materials).

\paragraph{Blender Dataset}
Table \ref{tab:blender} shows comparison of the evaluated methods on Blender test NeRF instances. Results show clear improvement on perceptual metrics for K-means quantization. The size of these models following compression with BZIP2 is noticeably larger for the K-means quantization. As BZIP2 compresses most effectively for repeated information, it is possible that the distribution of quantized weights is more uniform under the K-means clustering (and is therefore being used more completely). Note that the compressed size obtained under this process is very small, with the NeRF model compressed to a size of approximately 16kb. The qualitative evaluation shows that even under this extreme compression a clear signal is able to be reconstructed (see Figure \ref{fig:drums}). 

\begin{table}[]
\begin{tabular}{|lrrrr|}
\hline
\textbf{Method}  & \multicolumn{1}{l}{PSNR} & \multicolumn{1}{l}{SSIM} & \multicolumn{1}{l}{LPIPS} & \multicolumn{1}{l|}{Size (kb)} \\ \hline
Chair &&&&\\ \hline
K-means         & \textbf{29.78}    & \textbf{0.95}     & \textbf{0.08}      & 15.58   \\
Distributional         & 26.76    & 0.91     & 0.16      & 13.45   \\
Minmax         & 27.69    & 0.93     & 0.14      & 10.95   \\
Explicit {[}-1, 1{]}   & 28.40    & 0.94     & 0.12      & 14.49   \\ \hline
Drums&&&& \\ \hline
K-means       & \textbf{21.03}    & \textbf{0.81}     & \textbf{0.25}      & 15.85   \\
Distributional     & 19.72    & 0.72     & 0.41      & 13.40   \\
Minmax        & 19.35    & 0.71     & 0.42      & 11.63   \\
Explicit {[}-1, 1{]}   & 20.47    & 0.77     & 0.32      & 15.00   \\ \hline
Ficus&&&& \\ \hline 
K-means       & \textbf{23.85}    & \textbf{0.89}     & \textbf{0.12}      & 16.17   \\
Distributional       & 23.03    & 0.86     & 0.22      & 13.42   \\
Minmax          & 22.71    & 0.85     & 0.25      & 11.85   \\
Explicit {[}-1, 1{]}  & 23.39    & 0.88     & 0.16      & 14.10   \\ \hline
Hotdog&&&& \\ \hline 
K-means    & \textbf{27.52}    & \textbf{0.88}     & \textbf{0.19}      & 15.52   \\
Distributional     & 25.04    & 0.80     & 0.36      & 13.24   \\
Minmax     & 25.60    & 0.83     & 0.29      & 11.51   \\
Explicit {[}-1, 1{]}  & 26.07    & 0.83     & 0.30      & 14.83   \\ \hline

Lego&&&& \\ \hline 
K-means       & \textbf{22.88}    & \textbf{0.82}     & \textbf{0.16}      & 15.69   \\
Distributional       & 21.37    & 0.77     & 0.27      & 13.33   \\
Minmax         & 20.79    & 0.76     & 0.28      & 10.92   \\
Explicit {[}-1, 1{]}   & 21.89    & 0.79     & 0.21      & 14.89   \\ \hline
Materials \\ \hline 
K-means            & \textbf{21.51}    & \textbf{0.82}     & \textbf{0.23}      & 15.36   \\
Distributional            & 19.88    & 0.76     & 0.37      & 13.22   \\
Minmax           & 19.76    & 0.75     & 0.37      & 10.64   \\
Explicit {[}-1, 1{]}      & 20.77    & 0.78     & 0.34      & 14.03   \\ \hline
Mic&&&& \\ \hline
K-means        & \textbf{26.07}    & \textbf{0.93}     & \textbf{0.13}      & 16.02   \\
Distributional      & 23.68    & 0.89     & 0.24      & 13.43   \\
Minmax        & 23.09    & 0.89     & 0.24      & 11.30   \\
Explicit {[}-1, 1{]}    & 24.96    & 0.92     & 0.18      & 14.44   \\ \hline
Ship&&&& \\ \hline 
K-means       & \textbf{24.49}    & \textbf{0.69}     & \textbf{0.37}      & 15.87   \\
Distributional       & 23.45    & 0.65     & 0.46      & 13.35   \\
Minmax         & 23.48    & 0.66     & 0.45      & 11.08   \\
Explicit {[}-1, 1{]}   & 23.93    & 0.66     & 0.44      & 14.52   \\ \hline
\end{tabular}
\caption{Quantitative Results on Blender (NeRF), Architecture: 4 hidden layers, 64 units per layer, 3-bit quantization). Metrics averaged over test images.} 
\label{tab:blender}
\end{table}

\section{Discussion and Limitations}\label{sec:dis}


\paragraph{Periodic Repartitioning}  While recalculating partitions more frequently reduces quantization error (see Figure \ref{fig:repartition}), this overhead may make frequent repartitioning using K-means impractical for large architectures with high quantization resolution. The computational complexity of the implemented K-means algorithm is $O(mn +n\log n)$ where $m$ is the number of partitions and $n$ is the data points \cite{steinberg_kmeans1d_nodate}. For our case this depends on the bits-per-weight $k$, (as $m=2^k$); the number of neurons at each layer $w_{in}, w_{out}$ (as $n=w_{in}*w_{out}$); and the number of layers. As such, the clustering operation scales inefficiently with large architecture sizes and bits-per-weight. In practice, we find that repartitioning every 50 to 100 epochs to balance this cost in our experiments to a manageable overhead. To a lesser extent, the overhead for Distributional and Minmax is also affected by repartitioning frequency. While a balance between the increased computation cost and perceptual benefit of more frequent repartitioning is a subjective consideration for the experimenter, we note as a societal impact the high global energy consumption in machine learning \cite{garcia-martin_estimation_2019}.

%

\begin{figure}[h!]
    \centering
    \includegraphics[width=0.45\textwidth]{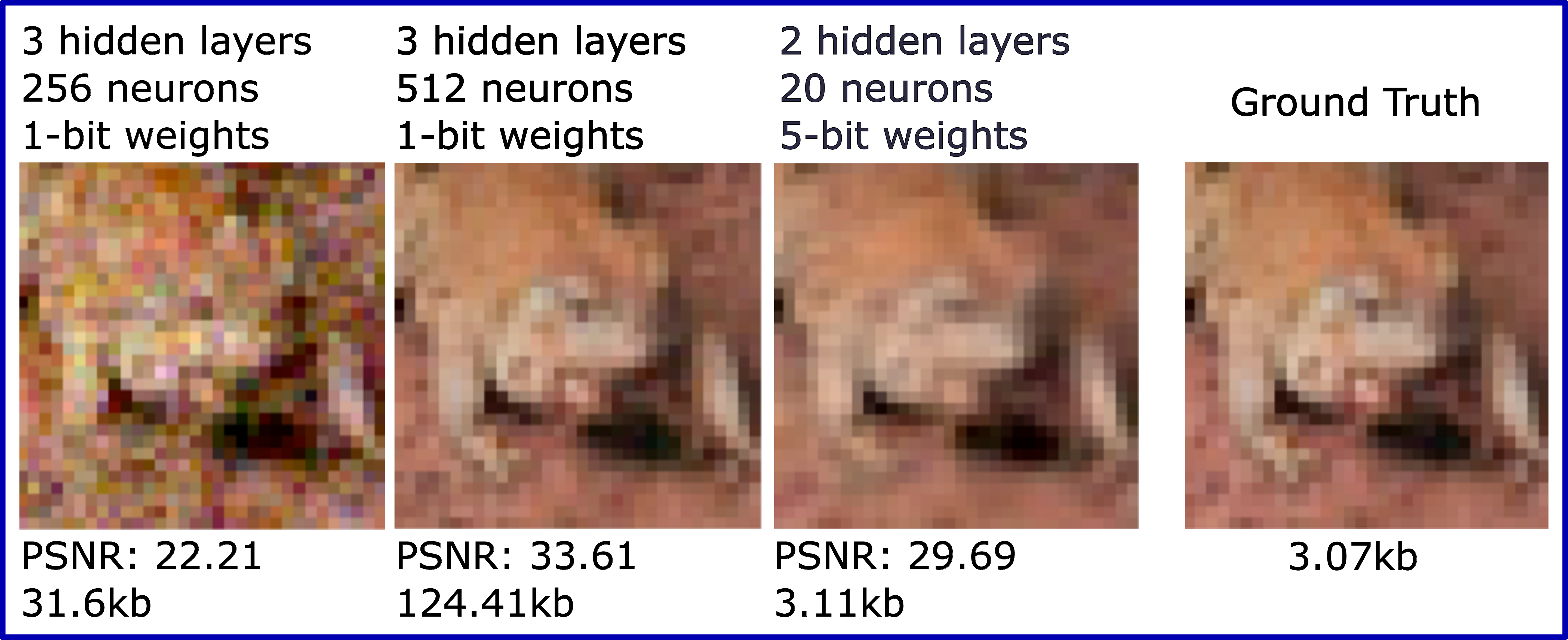}
    \caption{Trading network capacity for quantization levels. While it is possible to use K-means weight quantization to enable 1-bit reconstructions with sufficient network capacity, this comes at a trade-off for memory efficiency.}
     \label{fig:1bitreconstruction}
\end{figure}

\paragraph{Network Capacity and Quantization Trade-Off} One interesting experimental observation is an apparent trade-off between the network capacity and quantization. In Figure \ref{fig:div2ksurface}, we show the effect of weight quantization at different bit rates using K-means quantization. By considering the level-sets of the induced PSNR, we note that multiple configurations of network width, depth, and quantization level can lead to the same PSNR. The architectures differ however on the memory consumption, with exponential memory increase in the number of layers. A consequence of this is architectural mitigation of reconstruction failure even with extreme weight quantization. Figure \ref{fig:1bitreconstruction} shows this visually on CIFAR: a) shows a network of 3 hidden layers, 256 neurons per layer, 1-bit weights (PSNR: 22.21, 31.6kb); b) 3 hidden layers 512 neurons, 1-bit weights (PSNR: 33.61; 124.41kb). As compression, this is not very useful: the quantized network in b) is approximately 40x the original CIFAR image - however it clearly demonstrates the performance trade-off between weight quantization and neural architecture. As comparison, we note that improved memory efficiency can be obtained with a higher number of bits-per-weight as shown in: c) where we have 2 hidden layers, 20 neurons, quantized to 5-bit weights (PSNR:29.69, 3.11kb) due to the architecture memory cost. 

\paragraph{Future Work} Further improvements can additionally be applied to our chosen quantization schemes. In particular, the quantization methods employed are \textit{deterministic} which can introduce patterns and artefacts in the quantized mapping; the well-known method of \textit{dithering} is one way to avoid this issue \cite{gersho_vector_1992,salomon_data_2007}. To a certain extent recalculating centroids introduces a source of stochasticity which may help obviate this issue, but a formal evaluation remains an area for future work. As a side note, we find that using a log L2 loss function (i.e. directly optimising for PSNR) produced higher accuracy reconstructions at a given training epoch than the standard L2 loss function more common in the literature (see supplementary material), a result that held consistently across ablations of network architecture and activation function. We note that the logarithm is a monotonic operation, and therefore does not change the theoretical minimum of the converged network. As a result, it is interesting to see that this modification led to consistent improvements in the 2D evaluation. This result did not hold in general for other more expressive experiments, such as NeRF. Formal investigation to the cause of this observation remains a potential avenue of enquiry.

\section{Conclusion} 

We investigated the use of non-uniform quantization for the compression of implicit neural functions. By accounting for the weight distribution of the neural network layers, we are able to achieve higher performance at lower bits-per-weight than under uniform quantization approaches. We have additionally shown that there exists a trade-off between the network capacity and the weight quantization levels, with an extreme (binary) quantization able to be compensated with sufficient network capacity for simple experiments. Our method enables compression of neural radiance fields to a large degree compared to the original NeRF model. As the strategy involves a modification of the quantization strategy employed to the neural weights, it is possible that this may also be applied to other large implicit neural representations. Of particular interest is the potential to apply it to high resolution methods such as kiloNeRF, which have a large memory footprint of 100MB or SNeRG with 90MB \cite{reiser_kilonerf_2021, hedman_baking_2021}. As these methods are optimised for real-time inference this would be a step towards lightweight real-time NeRF models.

{\small
\bibliographystyle{ieee_fullname}
\bibliography{references}
}

\end{document}